\documentclass[a4paper,fleqn]{cas-dc}

\usepackage[square,numbers,sort&compress]{natbib}
\def\tsc#1{\csdef{#1}{\textsc{\lowercase{#1}}\xspace}}
\tsc{WGM}
\tsc{QE}

\usepackage[inline]{enumitem}
\usepackage{amsmath}
\usepackage{amssymb}

\usepackage{graphicx}

\usepackage{hyperref}
\usepackage{orcidlink}
\hypersetup{
 colorlinks,
 linkcolor={red},
 citecolor={blue},
 urlcolor={blue}
}

\setlist[itemize]{noitemsep, nosep, itemsep=3pt, topsep=3pt, itemindent = -1em}
\setlist[enumerate]{noitemsep, nosep, itemsep=3pt, topsep=3pt,itemindent = -1em}

\DeclareMathOperator*{\argmin}{argmin}

\begin{document}
\let\WriteBookmarks\relax
\def\floatpagepagefraction{1}
\def\textpagefraction{.001}

\shorttitle{Self-Supervised Calibration of Scientific Instruments Using Physical Consistency Constraints}
\shortauthors{M.~Rejmund, A.~Lemasson}

\title[mode = title]{Self-Supervised Calibration of Scientific Instruments Using Physical Consistency Constraints}


\author{M.~Rejmund\,\orcidlink{0009-0009-8626-8756}}[orcid = 0009-0009-8626-8756]
\ead[M. Rejmund: ]{mrejmund@ganil.fr}
\author{A.~Lemasson\,\orcidlink{0000-0002-9434-8520}}[orcid =0000-0002-9434-8520]
\affiliation{
organization = {GANIL, CEA/DRF - CNRS/IN2P3},
addressline= {Bd Henri Becquerel, BP 55027},
postcode= {F-14076},
city= {Caen Cedex 5},
country = {France}
}




\begin{abstract}
Calibration remains one of the principal obstacles to the deployment of machine learning in scientific 
instrumentation because it typically relies on expert intervention, dedicated procedures, and 
manually labelled data.

We introduce a physics-informed self-supervised framework that jointly learns latent detector 
calibration parameters and task-specific predictions directly from raw measurements without 
requiring pre-calibrated signals or external labels. The method exploits known physical 
constraints to generate pseudo-labels iteratively, transforming calibration into a self-supervised 
optimization problem.

The approach is demonstrated for ionic charge-state determination in the VAMOS++ magnetic 
spectrometer, where the calibration of a segmented ionization chamber and the inference of ionic 
charge states are learned simultaneously. Starting from a weak prior on the mean ionic charge state, 
the model progressively refines its predictions through iterative fractional pseudo-labelling driven by the 
discrete nature of atomic masses.

Beyond accurate ionic charge-state reconstruction, the inferred calibration coefficients provide a 
compact representation of the detector state that enables automated monitoring of gain drifts, 
pressure variations, and detector aging. The resulting labels can subsequently be transferred 
to specialized models that quantify detector imperfections and track their spatial and temporal evolution.

These results establish a general paradigm for self-calibrating and self-monitoring scientific 
instruments and represent a step toward intelligent experimental systems capable of autonomous 
calibration, analysis, and performance optimization.

\end{abstract}



\maketitle


\section{Introduction}

Modern scientific instruments generate increasingly large volumes of complex, high-dimensional data. From particle detectors 
and astronomical observatories to medical imaging devices and industrial sensor networks, extracting reliable information 
from these measurements relies critically on accurate detector calibration. Calibration procedures establish the relationship 
between raw detector signals and physically meaningful quantities, enabling quantitative interpretation of experimental observations.

Despite recent advances in machine learning, calibration remains one of the principal obstacles to the deployment of artificial 
intelligence in scientific instrumentation. Conventional calibration procedures are typically expert-driven, experiment-specific, 
and require dedicated measurements, iterative corrections, and extensive human intervention. In many applications, the generation 
of labelled training data itself depends on approximate calibration procedures, creating a circular dependency between calibration 
and inference. As a consequence, supervised learning approaches often suffer from limited scalability, reduced reproducibility, and 
poor adaptability to evolving experimental conditions.

These limitations have motivated increasing interest in self-supervised and physics-informed machine learning approaches, where prior 
knowledge is incorporated directly into the learning process. In scientific domains, physical laws, conservation principles, and consistency 
constraints can provide powerful sources of supervision, reducing or eliminating the need for manually labelled data. Rather than treating 
calibration as a prerequisite for data analysis, an alternative paradigm is to formulate calibration itself as a learning problem constrained 
by known physical relationships.

Such an approach enables a broader transition from static calibration procedures toward intelligent instrumentation. In this emerging paradigm, 
detector systems continuously infer their own calibration state from incoming data, monitor their performance over time, identify deviations 
from optimal operating conditions, and provide information for corrective actions. Calibration parameters are no longer viewed merely as 
intermediate quantities required for data analysis but become detector-state observables that characterize the health and performance 
of the instrument.

In this work, we introduce a physics-informed self-supervised framework that jointly learns detector calibration parameters and 
task-specific predictions directly from raw measurements without requiring pre-calibrated signals or externally provided labels. 
The approach combines weak physical priors, iterative fractional pseudo-labelling, and domain-specific consistency constraints to progressively 
refine both the detector calibration and the inferred quantities of interest.

We demonstrate the proposed framework using the VAMOS++ large-acceptance magnetic 
spectrometer~(\citet{Pullanhiotan2008, Rejmund2011}) at Grand Acc\'el\'erateur National d'Ions Lourds 
as a challenging case study. 
VAMOS++ is widely used in nuclear physics experiments for the identification of reaction products produced in heavy-ion collisions. 
Accurate reconstruction of the ionic charge state is essential for determining the atomic mass of detected ions. However, ionic 
charge-state identification depends critically on the calibration of a segmented ionization chamber, whose response may vary with 
detector gain, gas pressure, operating conditions, and local detector imperfections.

Previous machine-learning approaches applied to VAMOS++ have demonstrated substantial improvements in ion trajectory reconstruction 
and particle identification. Nevertheless, these methods relied on approximate online calibrations to generate partially labelled datasets. 
The requirement for labelled data remains a major limitation for the broader deployment of AI-based analysis pipelines.

The present work addresses this challenge by eliminating the dependence on pre-calibrated detector signals. Instead, detector calibration 
coefficients and ionic charge-state assignments are jointly inferred from raw detector observables using only weak physical priors 
and physically motivated consistency constraints. Specifically, the discrete nature of atomic masses is exploited to generate 
pseudo-labels iteratively, transforming detector calibration into a self-supervised optimization problem.

Beyond enabling automatic label generation, the inferred calibration coefficients provide a compact representation of the detector state 
and can therefore be monitored as a function of time, operating conditions, or detector coordinates. Variations in these coefficients 
may reveal gain drifts, pressure fluctuations, aging effects, or other changes in detector response. Furthermore, the automatically 
generated labels can be transferred to specialized models designed to account for detector non-uniformities and higher-order effects, 
enabling the spatial and temporal characterization of detector imperfections.

The present work makes four primary contributions:

\begin{enumerate}
\item We formulate detector calibration as a physics-informed self-supervised learning problem.

\item We introduce an iterative fractional pseudo-labelling framework that jointly estimates latent calibration 
parameters and task-specific predictions directly from raw measurements.

\item We demonstrate that physical consistency constraints can be used
to infer calibration parameters directly from raw measurements.

\item We establish a pathway toward intelligent instrumentation by integrating calibration, analysis, 
and detector diagnostics within a unified framework.
\end{enumerate}

Although demonstrated on a nuclear-physics detector, the proposed framework is applicable to a broad range of scientific instruments 
for which calibration parameters are latent variables and physical consistency constraints are available. Examples include particle detectors, 
astronomical instrumentation, medical imaging systems, fusion diagnostics, and industrial sensor networks.

The proposed framework shares conceptual similarities with expectation-maximization algorithms, 
self-training methods, and physics-informed machine learning, while addressing the unique challenge 
of learning latent calibration parameters directly from raw detector measurements.

These results establish a pathway toward self-calibrating and self-monitoring scientific instruments in which calibration, 
data analysis, and performance evaluation are integrated within a unified adaptive framework.

\section{Related Work}

The increasing complexity and data throughput of modern scientific instruments have stimulated 
growing interest in artificial intelligence methods for calibration, and data analysis.  
However, most existing approaches assume that detector calibration has already been performed and that 
labelled datasets are available for training.

\subsection{Machine Learning for Scientific Instrumentation}

Supervised learning methods have demonstrated remarkable performance for tasks such as event 
classification, particle identification, trajectory reconstruction, detector calibration, anomaly detection, 
and experimental control. In nuclear and particle physics, deep neural networks have been applied 
successfully to detector reconstruction pipelines, significantly improving processing speed and analysis 
reproducibility~(\citet{Radovic2018, Guest2018, Carleo2019, Albertsson2018}).

Nevertheless, the deployment of supervised learning in scientific instrumentation remains limited 
by the cost and availability of labelled data. In many cases, labels are obtained through dedicated 
calibration procedures, simulations, or expert-driven analyses. Consequently, the generation of 
training datasets often depends on the very calibration procedures that machine learning aims to replace.

This circular dependence creates a major bottleneck for the adoption of AI in scientific experiments, 
particularly in environments where detector responses evolve over time due to changing operating 
conditions or environmental variations.

\subsection{Weakly Supervised and Self-Supervised Learning}

Weakly supervised and self-supervised learning methods have emerged as promising alternatives 
in situations where fully labelled datasets are unavailable or expensive to obtain.

Pseudo-labelling, self-training, expectation-maximization methods, and consistency-based 
learning~(\citet{Lee2013, Xie2020, Sohn2020, VanEngelen2020, Demptster1977})
have been successfully employed across diverse domains to iteratively infer latent variables and 
improve prediction quality. In these approaches, a model initially trained using limited supervision 
generates labels for previously unlabelled samples, which are subsequently used to refine the model itself.

Recent work has demonstrated the potential of weakly supervised learning for detector calibration 
and particle identification in nuclear physics~(\citet{Rejmund2025b}). In particular, fractionally 
labelled datasets obtained from approximate online calibrations have enabled significant 
improvements in analysis speed and reproducibility.

However, existing approaches still depend on partially calibrated detector signals and externally 
generated labels. As a result, the calibration procedure remains separate from the learning process itself.

\subsection{Physics-Informed Machine Learning}

Physics-informed machine learning seeks to incorporate prior scientific knowledge directly into 
the learning process through conservation laws, governing equations, symmetries, or consistency 
constraints.

Physics-informed neural networks and related approaches have shown that embedding physical 
constraints into model architectures or objective functions can substantially reduce data requirements 
while improving robustness and interpretability~(\citet{Raissi2019, Karniadakis2021, Cuomo2022, Willard2022}).

In experimental sciences, many observables satisfy known physical relationships that can be exploited 
as alternative sources of supervision. Examples include conservation laws, discrete quantum numbers, 
kinematic constraints, and integer-valued observables.

These constraints provide a natural mechanism for transforming calibration and inference tasks 
into self-supervised optimization problems, enabling the extraction of meaningful information from unlabelled data.

Unlike conventional physics-informed neural networks, which incorporate governing differential equations 
into the loss function, the proposed framework addresses a distinct problem: the inference of latent calibration 
parameters from raw detector measurements using discrete physical consistency constraints.

The objective is not to solve a forward physical model but to recover hidden instrument parameters that maximize agreement with known physical properties of the observed system.

\subsection{Toward Intelligent Instrumentation}

Beyond improving data analysis workflows, AI methods increasingly enable a transition toward 
intelligent instrumentation~(\citet{Brunton2022}), 
in which calibration, monitoring, diagnosis, and optimization become integrated within adaptive feedback loops.

The integration of self-calibration, performance monitoring, and adaptive optimization
is consistent with the broader vision of autonomous experimentation and self-driving
laboratories emerging across multiple scientific disciplines~(\citet{Tabor2018, Stach2021, Ren2023}).

Recent developments in digital twins, surrogate models, reinforcement learning, and Bayesian 
optimization~(\citet{Rasheed2020, Edelen2016, Emma2018, Edelen2024,Roussel2024})
 have demonstrated the potential of AI-assisted operation for accelerators, beam-lines, 
and complex scientific facilities.

However, the majority of these approaches assume that detector systems provide calibrated and 
reliable measurements. The challenge of enabling instruments to infer and monitor their own 
calibration state remains largely unexplored.

Calibration parameters themselves constitute valuable observables that characterize detector health 
and performance. Their temporal evolution may reveal gain drifts, pressure fluctuations, aging effects, 
or localized detector defects. Consequently, self-calibrating systems represent a key enabling technology 
for intelligent instrumentation.

\subsection{Positioning of the Present Work}

The present work addresses the calibration bottleneck by formulating detector calibration as a 
physics-informed self-supervised learning problem.

Unlike our previous supervised and weakly supervised approaches, which assume calibrated detector signals 
and externally provided labels~(\citet{Rejmund2025a, Rejmund2025b}), the proposed framework jointly infers 
latent calibration parameters and task-specific predictions directly from raw measurements.

The method combines weak physical priors, iterative fractional pseudo-labelling, and physically motivated 
consistency constraints to generate labels automatically while simultaneously estimating detector 
calibration coefficients.

The inferred calibration coefficients constitute detector-state observables that enable 
continuous monitoring of instrument performance and provide a basis for the spatial 
and temporal characterization of detector imperfections.

Although demonstrated using the VAMOS++ magnetic spectrometer, the proposed framework is 
broadly applicable to scientific instruments for which calibration parameters are latent variables and 
physical consistency constraints are available.

\section{Problem Formulation}

Many scientific instruments rely on calibration procedures that transform raw detector signals into 
physically meaningful observables. Conventional machine learning approaches typically assume 
that this calibration has already been performed and that labelled training data are available. 
However, in many experimental settings, the calibration parameters are themselves unknown, 
time-dependent, and costly to estimate.

We consider the general problem of jointly inferring latent calibration parameters and task-specific 
quantities directly from raw measurements without requiring pre-calibrated data or externally provided 
labels.

Let
$$
\mathbf{x} = {x_1, x_2, \dots, x_N}
$$
denote the raw measurements acquired by an instrument, where the individual components may 
correspond to detector signals, sensor responses, or auxiliary observables. Let
$$
\mathbf{c} = {c_1, c_2, \dots, c_M}
$$
represent a set of unknown calibration parameters characterizing the detector response.

The objective is to estimate a target quantity
$$
y = f(\mathbf{x}, \mathbf{c}),
$$
where both the calibration parameters ($\mathbf{c}$) and the target variable ($y$) are unknown.

In many scientific applications, the target quantity satisfies known physical constraints that 
can be expressed as
$$
\mathcal{P}(y, \mathbf{x}) = 0,
$$
where ($\mathcal{P}$) represents a physics-based consistency relation derived from conservation laws, 
discrete observables, symmetries, or governing equations.

Rather than relying on labelled examples ($\mathbf{x}, y$), we exploit these constraints to formulate 
calibration as a self-supervised optimization problem. The calibration parameters are learned by 
minimizing a physics-informed objective function,
$$
\mathbf{c}^{*}
=
\argmin_{\mathbf{c}}
\mathcal{L}_{physics}
\mathcal{P}(f(\mathbf{x}, \mathbf{c}), \mathbf{x})
,
$$
where ($\mathcal{L}_{\mathrm{physics}}$) quantifies deviations from the expected physical behaviour.

The inferred calibration parameters are subsequently used to generate pseudo-labels that iteratively improve the estimation 
of the target variable. This creates a closed learning loop in which calibration and inference are refined simultaneously.

\subsection{Case Study: Ionic Charge-State Determination in Magnetic Spectrometer}

The target variable is the ionic charge state ($q$), which is required to determine the atomic mass 
number of detected ions. Magnetic spectrometer provides measurements of the mass-to-charge 
ratio ($A/q$) and the Lorentz factor ($\gamma$), 
while the kinetic energy is measured using a segmented ionization chamber composed of  
independent detection segments.

The total kinetic energy can be expressed as
\begin{equation}
E_{tot}
=
\sum_{k=0}^{n}
E_k,
\label{Eq:ETOT}
\end{equation}
where ($E_k$) denotes the energy deposited in segment ($k$).
Assuming relativistic kinematics, the ionic charge state satisfies
\begin{equation}
q
=
\frac{E_{tot}}
{u(\gamma - 1)(A/q)},
\label{Eq:q}
\end{equation}
where ($u$) denotes the atomic mass unit.

In practice, the individual detector segments require calibration. The measured energies are therefore represented as
\begin{equation}
E_k
=
c_k E_k^{raw},
$$
\label{Eq:ECalib}
\end{equation}
where ($E_k^{raw}$) are the raw detector signals and ($c_k$) are unknown calibration coefficients.
The calibration coefficients ($\mathbf{c}={c_k}$) constitute latent variables that must be inferred jointly 
with the ionic charge state ($q$).

The key observation underlying the proposed approach is that atomic mass numbers are discrete quantities. For a given charge-state estimate, the reconstructed mass number is obtained as
\begin{equation}
A = (A/q) \left\lfloor
q +0.5
\right\rfloor.
\label{Eq:A}
\end{equation}

Physically meaningful solutions must satisfy the constraint that the reconstructed mass numbers lie close to integer values.
This discrete consistency condition provides a source of self-supervision that enables simultaneous calibration and 
ionic charge-state determination without requiring externally labelled data.

The following sections describe how these physical constraints are incorporated into an iterative fractional 
pseudo-labelling framework to infer both the detector calibration coefficients and the ionic charge states directly 
from raw detector signals.

\section{Experimental System: VAMOS++}

The proposed framework is demonstrated using the VAMOS++ at GANIL.
The spectrometer combines a large angular and momentum acceptance with high-resolution particle 
identification capabilities, enabling studies of nuclear structure of rare isotopes and reaction mechanisms.
Achieving unambiguous particle identification requires the simultaneous determination 
of several physical quantities, including the particle trajectory, velocity, mass-to-charge ratio, 
atomic number, and ionic charge state.

Charged particles transmitted through the spectrometer are tracked using position-sensitive detectors, from which their trajectories are reconstructed to determine the magnetic rigidity ($B\rho$), which along with the time-of-flight measurements lead ($A/q$) and $\gamma$.

The residual kinetic energy is measured using a segmented ionization chamber located at the focal 
plane of the spectrometer. The detector consists of ten independent segments that record the energy 
deposited by each ion as it traverses the detector gas volume.
The total kinetic energy is obtained as the sum of the segment contributions (Eq.~\ref{Eq:ETOT}).
Combined with the measured values of ($A/q$) and $\gamma$, the total kinetic energy enables 
determination of the ionic charge state ($q$) through relativistic kinematics (Eq.~\ref{Eq:q}).

In practice, accurate ionic charge-state determination requires precise calibration of the 
individual ionization chamber segments. Variations in electronic gains, gas pressure, 
detector aging, and local non-uniformities introduce time-dependent changes in detector 
response. Consequently, the measured segment energies 
cannot be directly summed but must first be corrected using calibration coefficients ($c_k$) (Eq.~\ref{Eq:ECalib}).

Conventional calibration procedures rely on dedicated measurements, approximate online 
calibrations, and iterative offline analyses requiring substantial expert intervention. 
VAMOS++ therefore provides a representative and challenging benchmark for self-supervised 
calibration. The detector combines high-dimensional measurements, latent calibration parameters, 
evolving detector conditions, and strong physical constraints, making it an ideal testbed for 
the development of intelligent instrumentation.

The evolution of calibration coefficients over time provides direct information 
about the detector’s state. Consequently, the proposed framework simultaneously 
addresses two objectives: automatic label generation and continuous monitoring 
of detector performance.

Additional details regarding the VAMOS++ spectrometer and its detector systems can be found in Refs.~\citet{Pullanhiotan2008, Rejmund2011, Rejmund2025b}.

\section{Physics-Informed Self-Supervised Calibration}

The proposed framework formulates detector calibration as a self-supervised learning problem 
in which latent calibration parameters and task-specific predictions are jointly inferred from raw 
detector measurements.

Rather than relying on externally provided labels or pre-calibrated detector signals, the method 
exploits physical consistency constraints to generate pseudo-labels iteratively. The resulting 
workflow combines weak physical priors, iterative self-training, and detector-state monitoring 
within a unified framework.

\subsection{Calibration Parameterization}

The segmented ionization chamber provides ten raw energy measurements,
$
E^{raw}
=
E_0^{raw},
\dots,
E_{9}^{raw}.
$
The calibrated total energy is modeled as given in Eq.~\ref{Eq:ECalib}
where
$
\mathbf{c}
=
{c_0,\dots,c_{9}}
$
are latent calibration coefficients.
The calibration coefficients are treated as trainable model parameters and optimized jointly with 
the ionic charge-state predictions.

To ensure physically meaningful solutions and improve optimization stability, the coefficients are 
constrained to predefined intervals,
$$
c_k
=
c_{\min}
+
\sigma(z_k)
(c_{\max}-c_{\min}),
$$
where ($z_k$) are unconstrained trainable parameters and ($\sigma(\cdot)$) denotes the sigmoid function.
This parameterization incorporates prior knowledge regarding the expected detector response directly 
into the model architecture and prevents unphysical calibration solutions.
The estimated charge state is subsequently obtained from Eq.~\ref{Eq:q}.

The network architecture therefore acts as a differentiable calibration model that maps raw detector 
signals to physically meaningful quantities through a small set of latent calibration parameters.

\subsection{Warm-Up Initialization with Weak Physical Priors}

Self-supervised optimization requires an initial estimate of the ionic charge state to avoid convergence 
toward unphysical solutions.
For the VAMOS++ datasets considered in this work~\cite{DataE826}, the average ionic charge state can be 
approximated from empirical systematics and online monitoring information.

During an initial warm-up phase, the model is trained for a fixed number of epochs using the loss function
$$
\mathcal{L}_{init}
=
\mathrm{RMSD}
\left(
q_{pred}
-
\bar{q}
\right),
$$
where ($\bar{q}$) denotes the estimated mean charge state and
RMSD stands for Root-Mean-Square-Deviation.
This weak prior provides only coarse global information and does not require event-level labels.

The objective of the warm-up stage is not to determine the final charge states accurately but 
rather to initialize the calibration coefficients within a physically plausible region of the parameter space.

\subsection{Physical Consistency Constraints}

The proposed framework exploits the discrete nature of atomic masses to generate self-supervision.
For a given charge-state prediction, the reconstructed mass number is obtained as
$$
A
=
(A/q)
\left\lfloor
q_{pred} + 0.5
\right\rfloor.
$$
Physically meaningful solutions require the reconstructed masses to cluster around integer values.
To account for possible systematic offsets in the predicted charge state, candidate charge-state corrections
$\Delta q \in [-3,3]$
are considered.
For each candidate shift, the reconstructed mass number is computed as
$$
A(\Delta q)
=
(A/q)
\left\lfloor
q_{pred}
+
\Delta q
+
0.5
\right\rfloor.
$$
The consistency of the reconstructed masses is quantified using
$$
\sigma(\Delta q)
=
\mathrm{RMSD}
\left(
A(\Delta q)
-
\left\lfloor
A(\Delta q)+0.5
\right\rfloor
\right).
$$
The optimal charge-state correction is then determined as
$$
\Delta q^\star
=
\argmin_{\Delta q}
\sigma(\Delta q).
$$
The corrected charge-state estimate becomes
$$
q_{label}
=
q_{pred}
+
\Delta q^\star.
$$

This procedure transforms a known physical property—the discrete nature of atomic masses—into 
a source of supervision.

The question might arise: \emph{Why optimize atomic mass consistency rather than atomic mass directly?}

An important distinction should be made between the optimization objective and the 
supervisory signal used during training.

The primary objective of the proposed framework is to determine the calibration coefficients that 
maximize the ionic charge-state resolution. From Eq.~\ref{Eq:q}, the charge-state resolution is 
directly related to the resolution of the calibrated total energy.

The reconstructed atomic mass number is obtained from Eq.~\ref{Eq:A}.
For a given event, the mass resolution is therefore dominated by the measured mass-to-charge ratio 
($A/q$), while the role of the inferred charge state is to assign the event to the correct integer 
ionic charge-state branch.

Because the experimental dataset contains a broad distribution of ionic charge states, 
an incorrect ionic charge-state assignment systematically shifts the reconstructed masses. As events 
originating from different true charge states are combined, the corresponding mass peaks no longer align, 
resulting in an increased apparent mass width.

Consequently, minimizing
$\mathrm{RMSD}\left(A-\lfloor A+0.5\rfloor\right)$
does not directly optimize the mass resolution itself. Instead, it provides a sensitive measure of 
charge-state assignment quality: the narrowest reconstructed mass peaks are obtained only when 
the correct ionic charge-state labels are assigned consistently across the full ionic charge-state distribution.

The integer-mass consistency criterion therefore serves as an indirect but physically meaningful 
supervisory signal for optimizing the detector calibration and ionic charge-state reconstruction.

\subsection{Iterative Fractional Pseudo-Labelling}

Pseudo-labels are generated only for events satisfying
$$
\left|
q_{label}
-
\left\lfloor
q_{label}+0.5
\right\rfloor
\right|
<
0.4.
$$
To ensure balanced coverage of the experimental phase space, pseudo-labelled events are 
sampled, for a given cycle, randomly from one of the predefined regions in the ($q_{label},A/q$) plane:
\begin{itemize}
\item $(q_{label} \leq 39 , (A/q) \leq 3)$,
\item $(q_{label} >39 , (A/q) \leq 3)$,
\item $(q_{label} \leq 39, (A/q) >3)$,
\item $(q_{label} >39 , (A/q) >3)$.
\end{itemize}
Events assigned pseudo-labels use
$$
q_{target}
=
q_{label},
$$
while all remaining events use
$$
q_{target}
=
\left\lfloor
q_{pred}+0.5
\right\rfloor.
$$
The model is subsequently retrained using the updated targets
$$
\mathcal{L}
=
\mathrm{RMSD}
\left(
q_{pred}
-
q_{rtarget}
\right).
$$

After each fractional pseudo-labelling cycle, all trainable parameters are reinitialized before retraining.
Complete network reinitialization after each pseudo-labelling cycle is a key component of the proposed framework.
Without reinitialization, inaccurate early pseudo-labels may become encoded in the network weights and reinforce 
themselves during subsequent iterations. Reinitialization prevents such confirmation bias by ensuring that only 
the updated pseudo-labels are propagated between training cycles.

The iterative pseudo-labelling procedure bears a strong conceptual resemblance to 
expectation-maximization algorithms~(\citet{Demptster1977}). 
During the pseudo-labelling phase, the current estimate of the latent calibration parameters 
is used to infer discrete hidden states in the form of ionic charge-state labels. These labels are subsequently 
held fixed while the network parameters are reinitialized and optimized. Although no explicit probabilistic 
model is assumed, the alternating estimation and optimization steps are analogous to the expectation 
and maximization stages of expectation-maximization-type algorithms.

Training and fractional pseudo-labelling cycles are repeated until convergence.

\begin{figure}
\centering
\includegraphics[width=1\columnwidth]{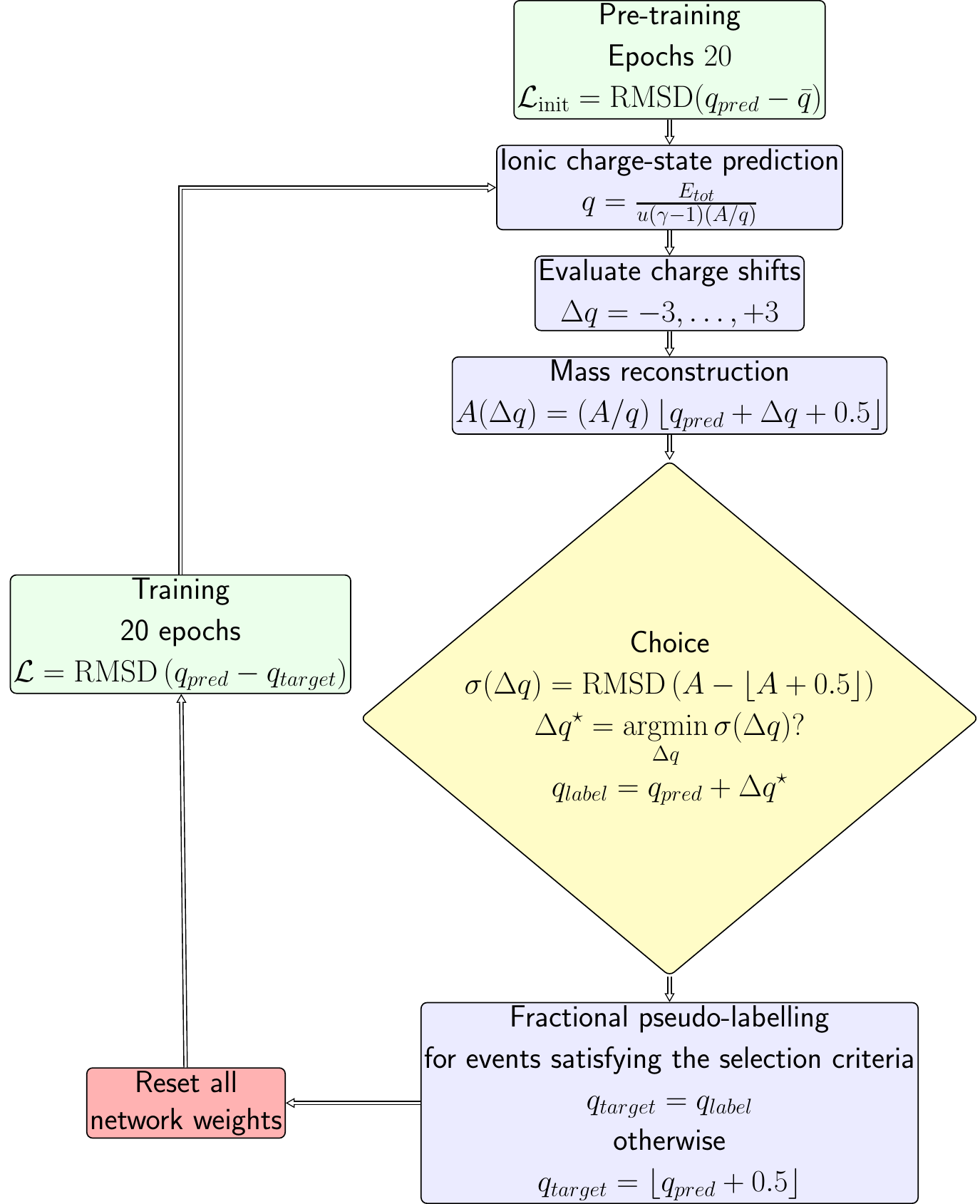}
\caption{{\bf Physics-informed self-supervised calibration workflow:}
Weak physical priors initialize the model, after which iterative fractional
pseudo-labelling driven by physical consistency constraints jointly
refines the detector calibration coefficients and ionic charge-state
predictions. The inferred calibration parameters can subsequently 
be monitored to characterize detector performance.
}
\label{fig:workflow}
\end{figure}

Figure~\ref{fig:workflow} illustrates the workflow of the proposed physics-informed 
self-supervised calibration.

\subsection{Calibration Coefficients as Detector-State Observables}

An important outcome of the proposed framework is that the inferred calibration 
coefficients constitute physically meaningful observables describing the detector state.
Unlike conventional calibration procedures, which typically generate static correction factors 
for a single analysis campaign, the present approach enables calibration parameters to be estimated 
repeatedly throughout data acquisition.
Monitoring the temporal evolution of the coefficients
$
\mathbf{c}(t)
$
provides a direct mechanism for identifying gain drifts, pressure fluctuations, detector aging, or changes in operating conditions.

Furthermore, the pseudo-labels generated by the present framework can be transferred to specialized supervised models designed to account explicitly for detector non-uniformities and higher-order effects.
Comparisons between the self-supervised calibration model and these specialized models enable detector imperfections to be quantified and localized as functions of detector coordinates and time.

Consequently, the proposed framework naturally separates global detector calibration from localized detector defects, establishing a foundation for self-calibrating and self-monitoring instrumentation.

\subsection{Implementation Details \label{sec:implementation}}

The performance and convergence stability of the self-supervised learning framework were 
evaluated using an experimental dataset containing $1\times10^7$ events~\cite{DataE826}. 
The dataset was 
partitioned into a training set ($80~\%$) and an independent validation set ($20~\%$) to 
monitor for potential overfitting. Training was executed in batches of $5\times 10^3$
events, with a complete dataset reshuffling performed at the start of each epoch.
The duration of a single training epoch was 
approximately $2$~s. The training was terminated when RMSD($q$), RMSD($A$), as well
calibration coefficients ($c_k$) reached the plateau. Remarkably, the training convergence was 
typically achieved within approximately 6 minutes.

\begin{figure*}
\centering
\includegraphics[width=0.49\textwidth]{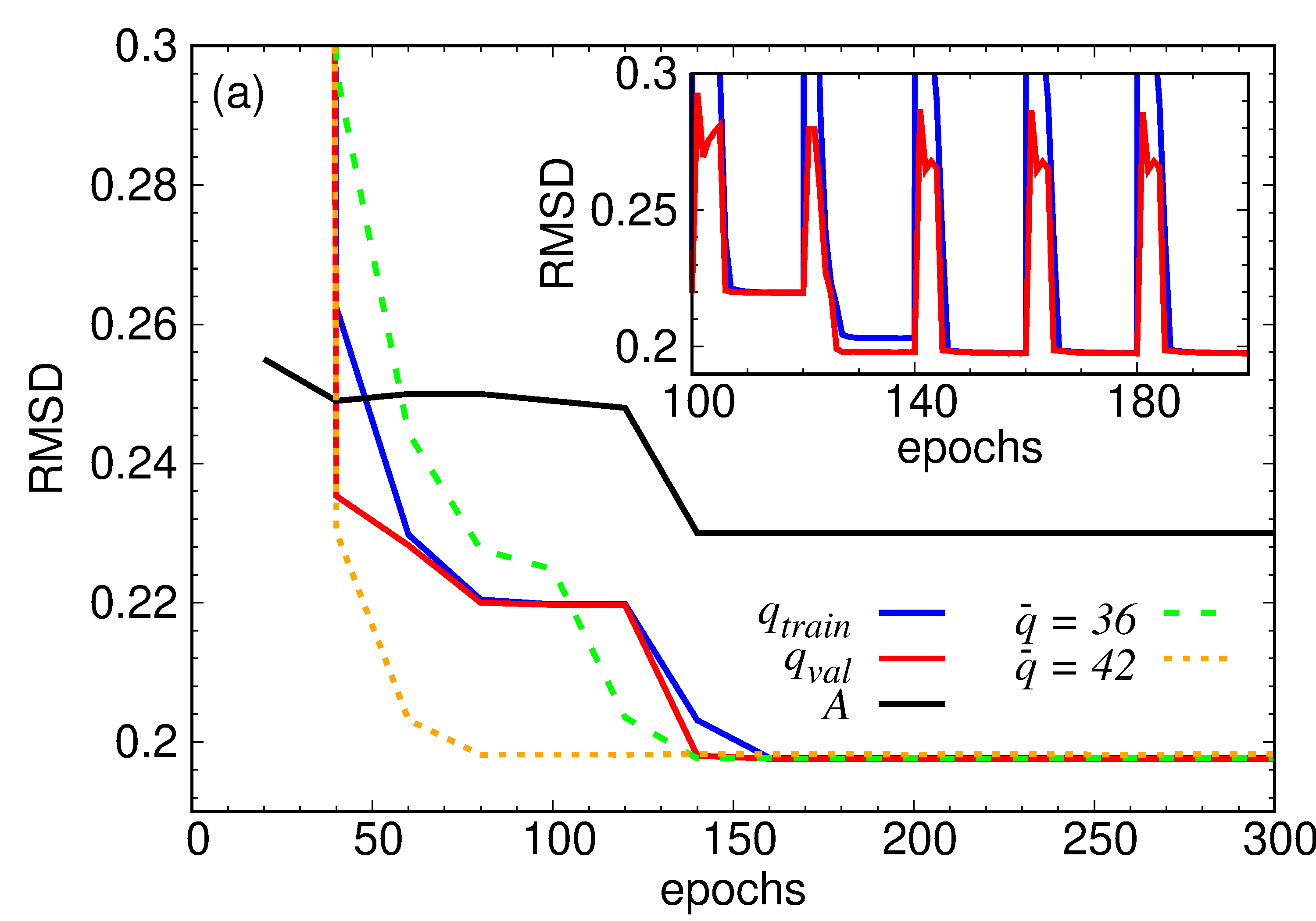}
\includegraphics[width=0.49\textwidth]{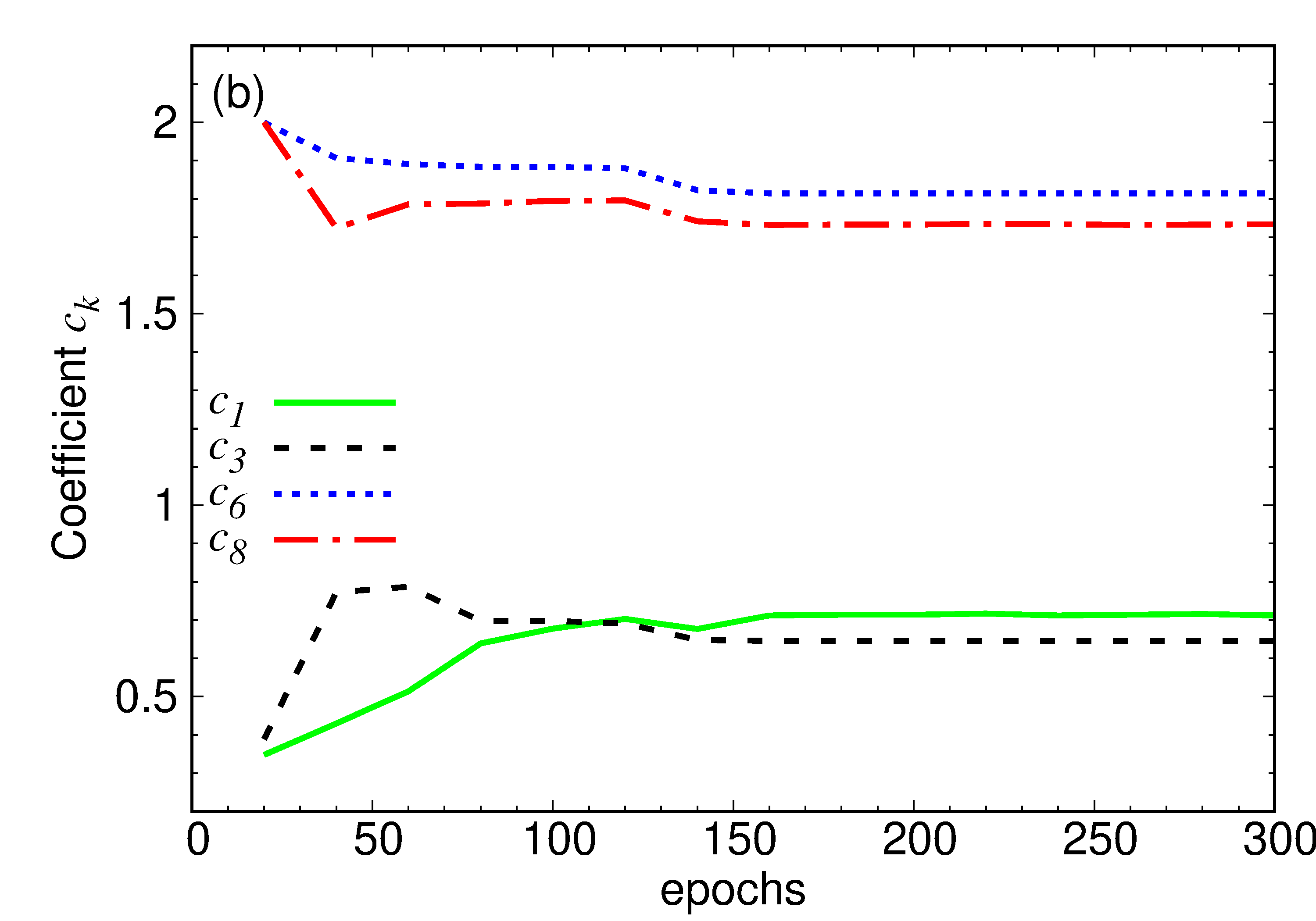}
\caption{{\bf Training convergence:} 
(a) The evolution of the RMSD($q$) for the training and validation, for initialization constraint $\bar q = 39$,
obtained for the epoch before fractional pseudo-labelling and RMSD($A$) obtained during the 
pseudo-labelling as a function of the epoch number. The dashed $(\bar q = 36)$ and dotted 
$(\bar q = 36)$ lines indicate
the training convergence of the model initialized with $\bar q = 36$ and $\bar q = 42$, 
respectively.
In the inset the RMSD($q$) is shown epoch-by-epoch 
for the epoch in range of $100-200$. (b) The evolution of several calibration coefficients ($c_k$)
evaluated during the fractional pseudo-labelling as a function of the epoch number. All coefficients have been 
normalized by a factor of $0.027$~MeV/ch.
}
\label{fig:trainingconvergence}
\end{figure*}

\section{Experimental Evaluation}

The proposed framework is evaluated using experimental data acquired with the VAMOS++ magnetic 
spectrometer. The objectives of the evaluation are to assess the convergence of the self-supervised 
calibration procedure, quantify the quality of the reconstructed ionic charge-state and mass distributions, 
validate the generated labels against a specialized network, and demonstrate the use of the inferred 
calibration coefficients as detector-state observables.

Unless stated otherwise, all results are obtained using the training configuration described in 
Section~\ref{sec:implementation}.

\subsection{Convergence of the Self-Supervised Calibration}

The first question addressed is whether the iterative fractional pseudo-labelling procedure converges toward a 
physically meaningful solution.

Figure~\ref{fig:trainingconvergence}(a) shows the evolution of the training and validation RMSD($q$) together with the 
RMSD($A$) evaluated during the pseudo-labelling stage. During the initial warm-up phase, the model is 
guided only by the weak prior on the average ionic charge state, $\bar q = 39$. Once pseudo-labelling is activated, 
the optimization progressively shifts from the initialization constraint toward the physically motivated 
integer-mass consistency criterion.

The RMSD($A$) decreases during the first fractional pseudo-labelling cycles before reaching a stable plateau, 
indicating convergence of the pseudo-label generation process. Simultaneously, the training (RMSD($q_{train}$)) 
and validation RMSD($q_{val}$) exhibit consistent behaviour, demonstrating that the iterative refinement procedure 
does not lead to overfitting. The system achieves a stable  convergence plateau at 
$\mathrm{RMSD}(q) = 0.198$. To contextualize this performance, this baseline closely approaches 
the $\mathrm{RMSD}(q) = 0.177$ achieved by the benchmark fully supervised specialized network~(\citet{Rejmund2025b}), which explicitly accounts for localized detector imperfections 
topologies. Concurrently, the atomic mass resolution metric, RMSD($A$), stabilizes at an optimal 
value of $0.23$.

The epoch-by-epoch tracking shown in the inset of Fig.~\ref{fig:trainingconvergence}(a) reveals narrow, 
periodic spikes that coincide exactly with the trainable parameter reinitialization. 
The rapid decay of these transients, demonstrates that the underlying pseudo-labels provide a 
highly robust and unambiguous optimization landscape, allowing the reinitialized network 
to swiftly recapture the global system characteristics without carrying over historical optimization biases.

The robustness of the method with respect to the initialization prior is illustrated by the additional 
training sequences performed with $\bar q = 36$ and $\bar q = 42$. Despite the substantially different 
initial conditions, all configurations converge toward same solutions, inferred calibration coefficients 
differing by less than $0.5~\%$ across all test cases. The model’s invariant convergence
indicates that the final calibration is driven primarily by the physical consistency constraints rather than the initialization prior and providing strong evidence for the robustness of the proposed self-supervised objective.

Figure~\ref{fig:trainingconvergence}(b) presents the evolution of representative calibration coefficients $c_k$ during training.  After an initial adjustment period, the coefficients stabilize,
providing further evidence that the optimization converges toward a consistent detector 
calibration.

Importantly, these coefficients maybe monitored to identify  changes in detector operating conditions.

\begin{figure*}
\centering
\includegraphics[width=0.49\textwidth]{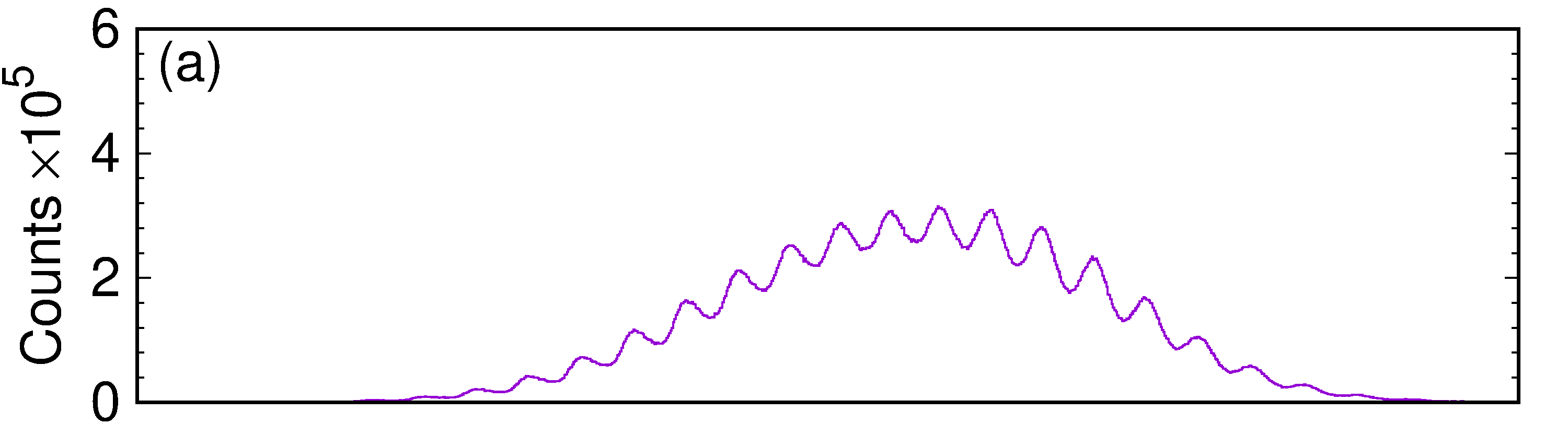}
\includegraphics[width=0.49\textwidth]{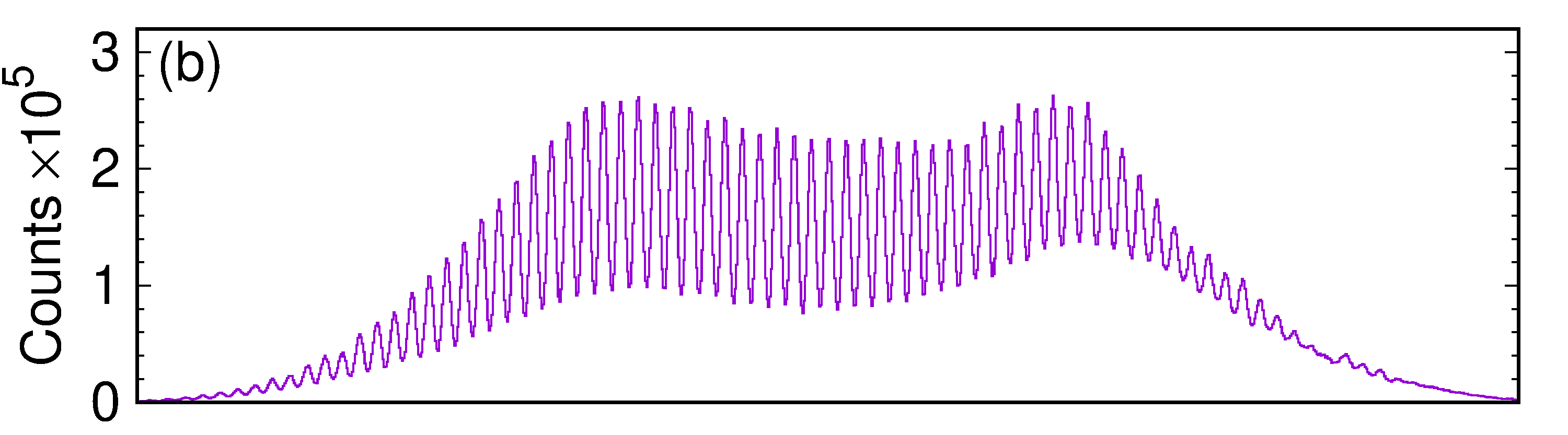}
\includegraphics[width=0.49\textwidth]{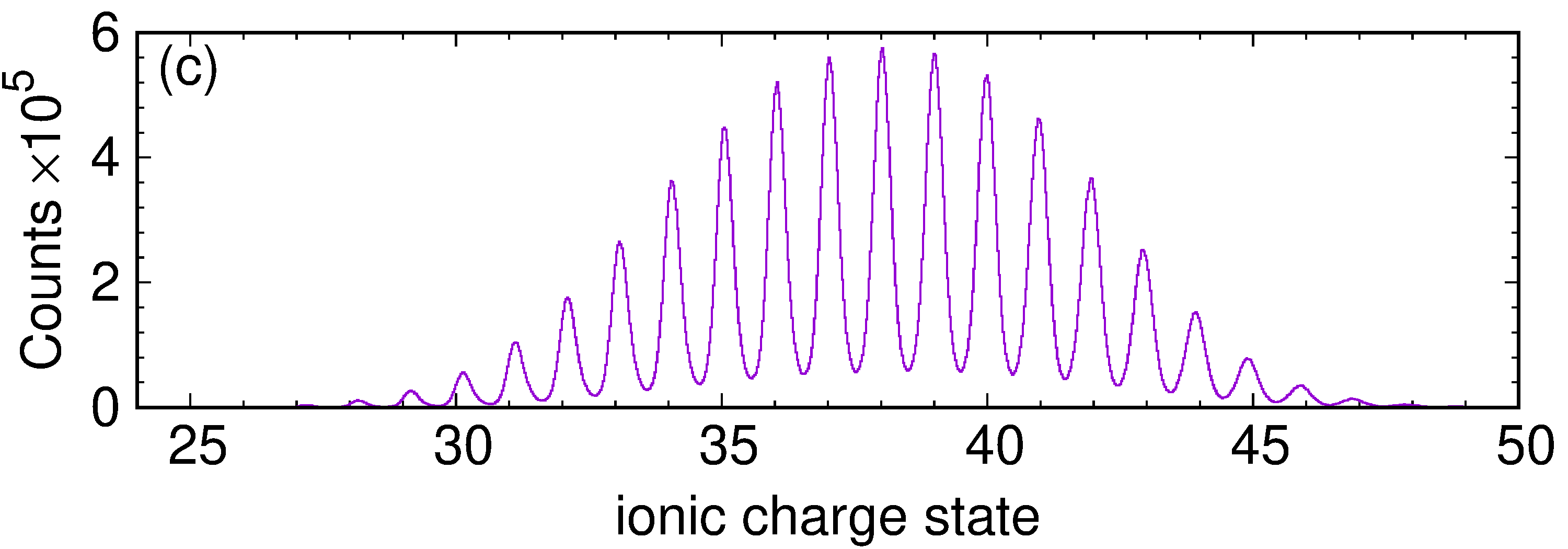}
\includegraphics[width=0.49\textwidth]{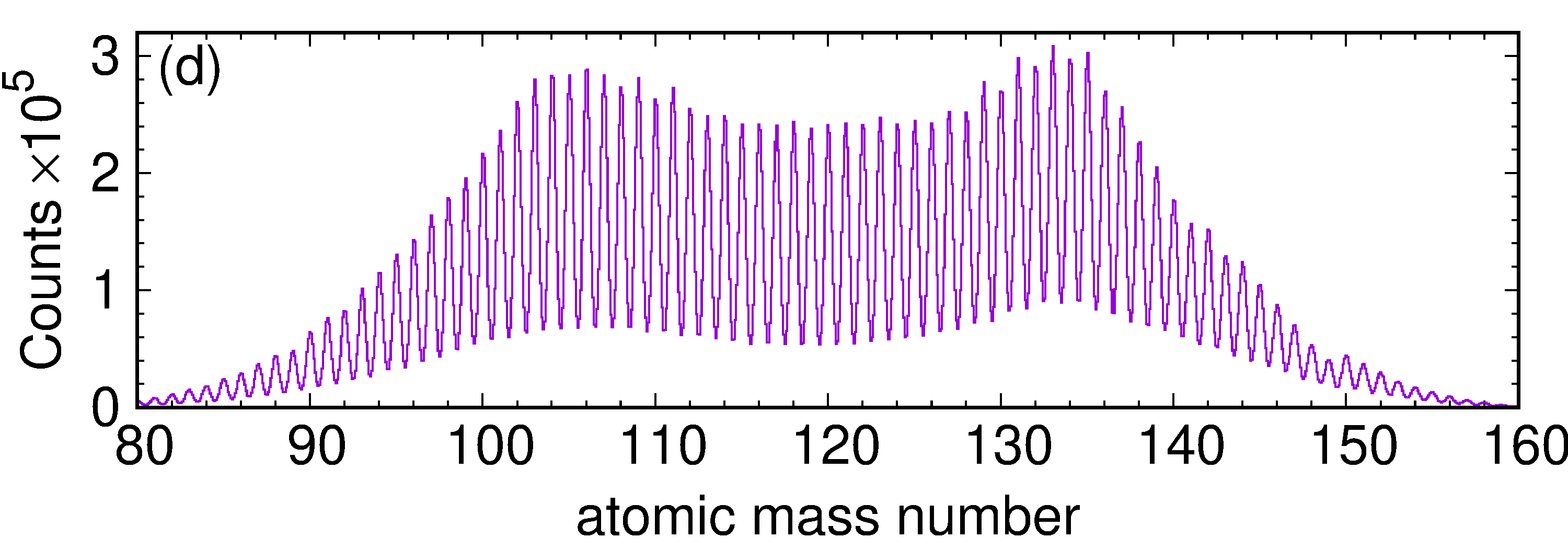}
\caption{{\bf Training progression:}
Evolution of the ionic charge-state and atomic mass spectra
(a) and (b) pre-training,
(c) and (d) convergence.
}
\label{fig:trainingprogress}
\end{figure*}

\subsection{Emergence of Ionic Charge-State and Mass Structure}

Figure~\ref{fig:trainingprogress} illustrates the evolution of the ionic charge-state and reconstructed mass 
spectra throughout training.
During the pre-training stage (Fig.~\ref{fig:trainingprogress}(a) and (b)), the weak prior on the average charge 
state provides only coarse guidance, resulting in broad charge-state distributions and poorly 
resolved mass peaks.
As the iterative fractional pseudo-labelling procedure progresses, the calibration coefficients and charge-state 
estimates are refined simultaneously. At convergence (Fig.~\ref{fig:trainingprogress}(c) and (d)), distinct ionic 
charge-state peaks emerge and the reconstructed masses exhibit a high resolution.

These results demonstrate that the proposed framework successfully transforms physical 
consistency constraints into an effective source of supervision. Starting from raw detector signals 
and a weak prior on the average charge state, the model autonomously discovers the calibration 
parameters required for accurate particle identification.

\begin{figure}
\centering
\includegraphics[width=1\columnwidth]{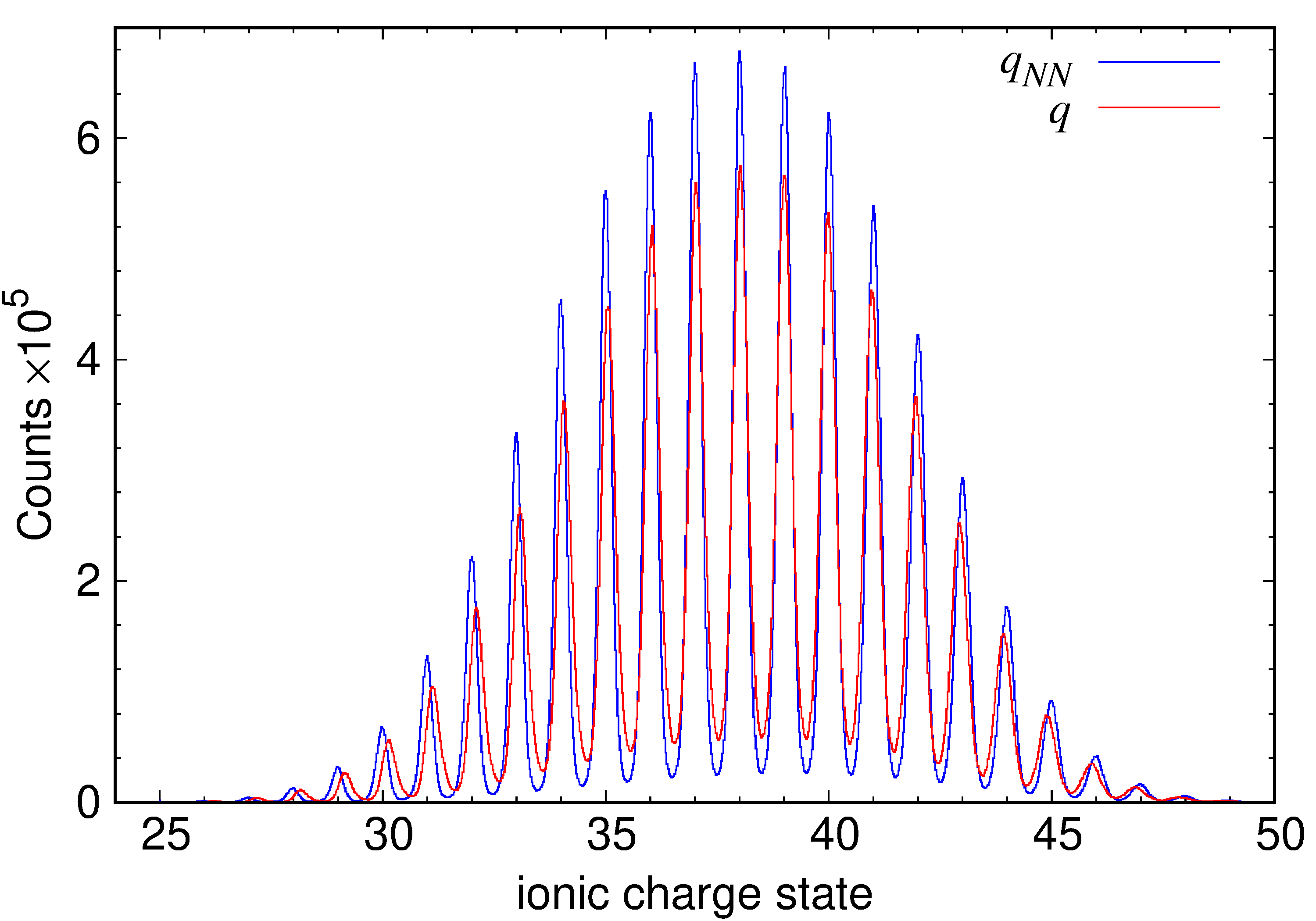}
\caption{{\bf Comparison with specialized network:}
The spectrum of the ionic charge state ($q$) obtained using the calibration coefficients 
($c_k$) obtained with the proposed method. It is compared to the spectrum ($q_{NN}$) obtained using
specialized network that accounts for the detection imperfections~(\citet{Rejmund2025b}).
}
\label{fig:qqnn}
\end{figure}

\subsection{Comparison with a Specialized Supervised Network}

To assess the quality of the automatically generated labels, the reconstructed ionic charge-state 
spectrum is compared with the output of the specialized network, using fractional labelling
introduced in~(\citet{Rejmund2025b}), 
which explicitly accounts for detector non-uniformities and localized imperfections.

As shown in Fig.~\ref{fig:qqnn}, the ionic charge-state distribution obtained using the proposed 
self-supervised calibration closely reproduces the spectrum reconstructed by the specialized network.
The supervised approach achieves a full-width at half-maximum (FWHM) resolution ranging from 
$1.0~\%$ for the highest atomic charge states to $1.2~\%$ for the lowest atomic charge states. 
In comparison, proposed self-supervised framework yields an exceptionally close resolution envelope 
of $1.1~\%$ to $1.3~\%$, respectively.

The summary of the self-supervised calibration performance is given in Table~\ref{tab:summaryperformance}.

This agreement demonstrates that the proposed method successfully eliminates the need for 
approximate online calibration while maintaining a level of performance comparable to that of 
a dedicated supervised approach. The differences are directly related with the detection imperfections.

Thus, the automatically generated labels produced by the self-supervised framework can be 
transferred directly to the specialized network, combining autonomous calibration with 
high-precision correction of detector-specific effects.

The proposed approach therefore establishes a hierarchical workflow in which self-supervised 
learning provides absolute calibration and label generation, while specialized supervised models 
capture higher-order detector effects.

\begin{table}[ht!]
\label{tab:summaryperformance}
\caption{Summary of the self-supervised calibration performance.}
\begin{tabular}{lc}
\hline
Metric & Value \\
\hline
Final RMSD($q$) & 0.198 \\
Final RMSD($A$) & 0.23 \\
Specialized network RMSD($q$) & 0.177 \\
Events used & $10^7$ \\
Batch size & $5\times10^3$ \\
Pseudo-labelling interval & 20 epochs \\
Pseudo-labelled fraction & 1/4 \\
Number of pseudo-labelling cycles & 8 \\
Training time & 6 m \\
\hline
\end{tabular}
\end{table}

\begin{figure}
\centering
\includegraphics[width=1\columnwidth]{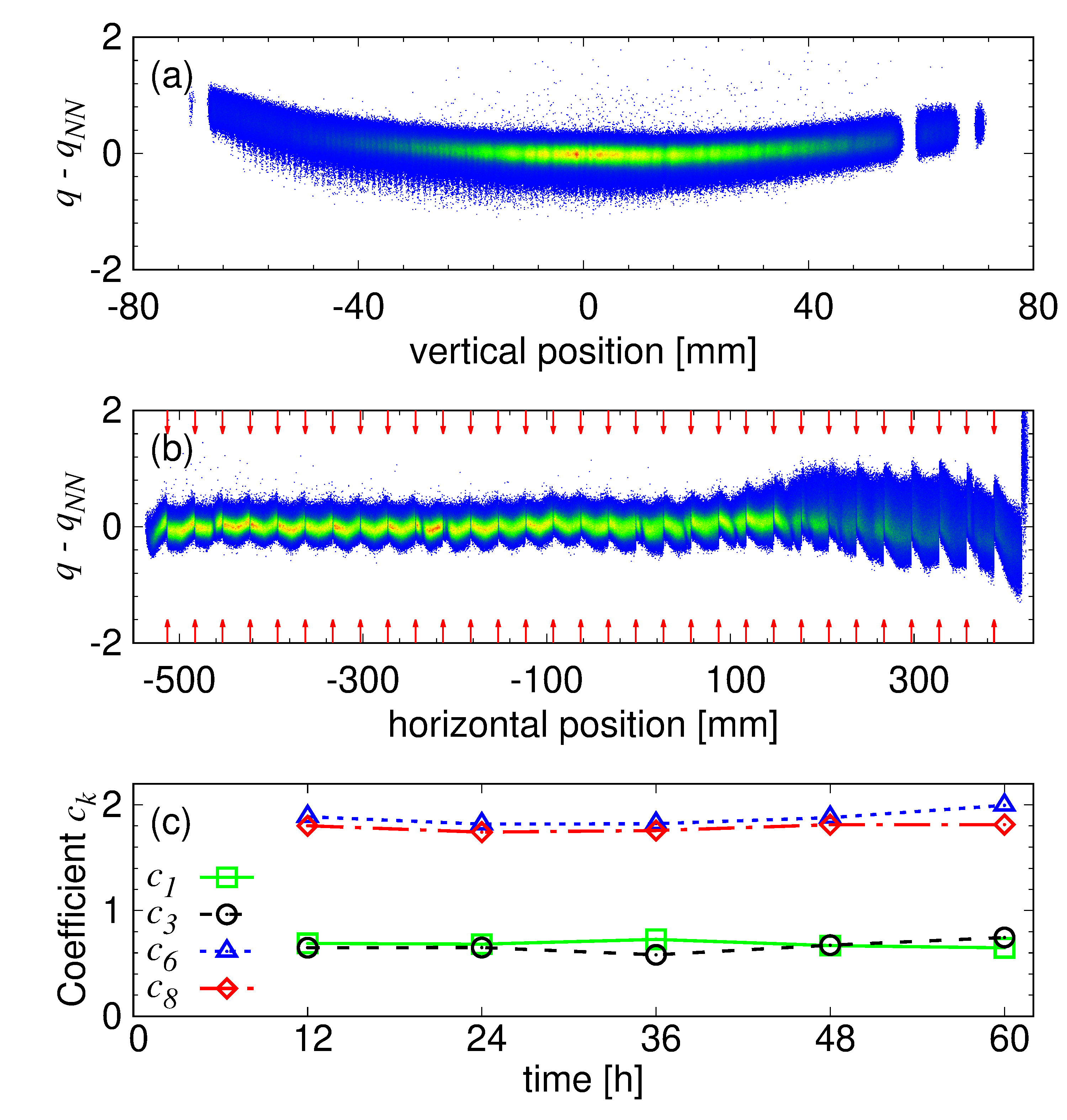}
\caption{{\bf Tracking detection imperfections:}
Two-dimensional correlation plot of the difference between ionic charge state ($q$) this work
and ($q_{NN}$) specialized network~(\citet{Rejmund2025b}), as a function of the (a) vertical and (b) horizontal 
position on the  entrance window of the ionization chamber. The red arrows in panel (b) indicate the positions 
of the vertical window holding wires, spaced by $30$~mm.
Panel (c): the evolution of several calibration coefficients ($c_k$) as a function of time, evaluated 
every $12$ hours. 
All coefficients have been normalized by a factor of $0.027$~MeV/ch.
}
\label{fig:detectorimperfections}
\end{figure}

\begin{figure}
\centering
\includegraphics[width=1\columnwidth]{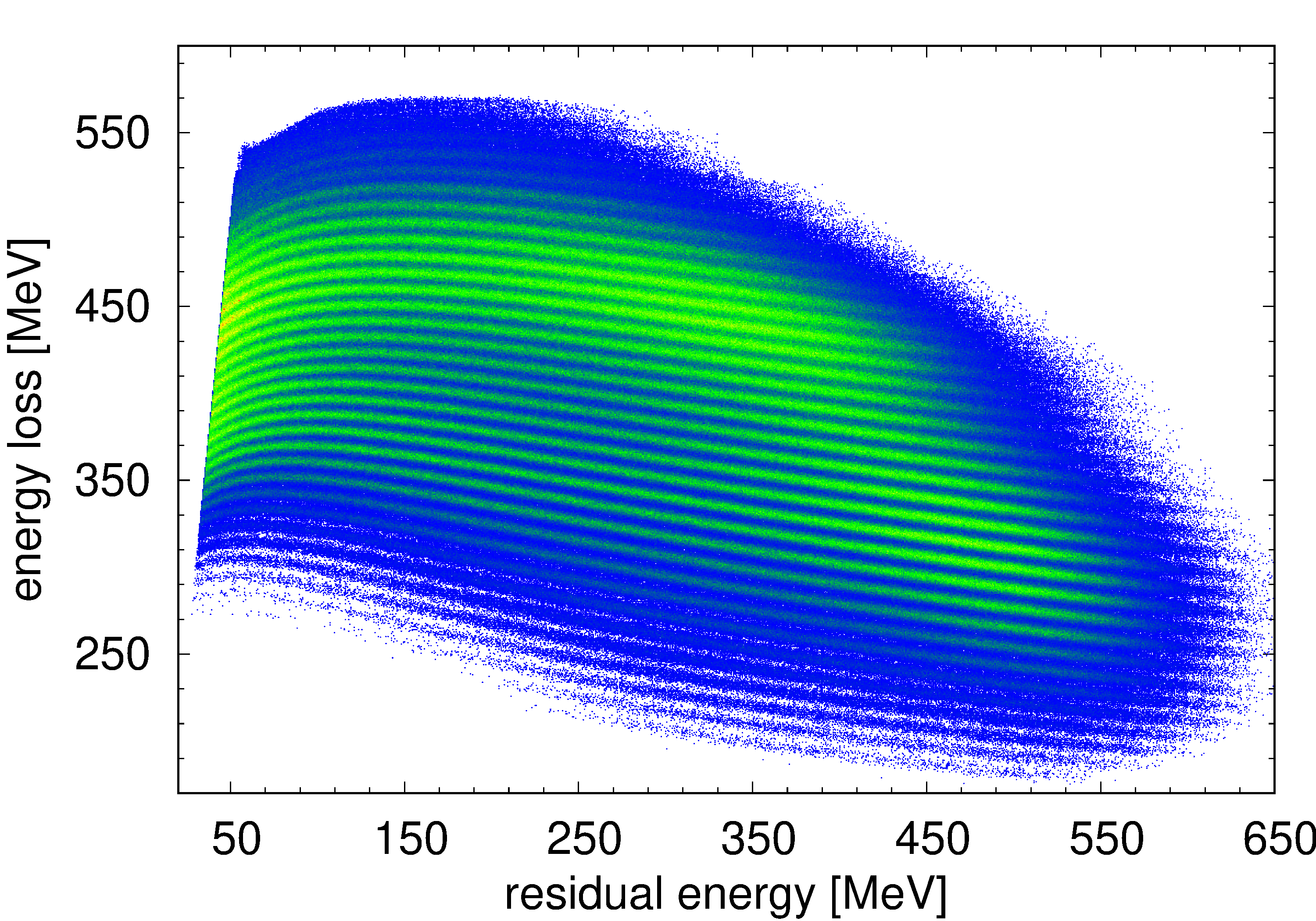}
\caption{{\bf Atomic number identification chart:}
Two-dimensional energy correlation plot between the energy loss ($\Delta E = \sum_{k=0}^{4} c_k E_k^{raw}$) 
and residual energy ($\Delta E_{res} = \sum_{k=5}^{9} c_k E_k^{raw}$)
using the calibration coefficients ($c_k$) obtained with the proposed
method.
The correlated bands correspond to different atomic numbers.
}
\label{fig:atomicnumber}
\end{figure}

\subsection{Tracking Detector Imperfections and Detector-State Evolution}

Beyond enabling automatic label generation, the inferred calibration coefficients provide direct 
information about the detector state.

Figure~\ref{fig:detectorimperfections}(a) and \ref{fig:detectorimperfections}(b) present the difference between the ionic charge states reconstructed 
by the proposed method and by the specialized network as a function of the vertical and 
horizontal coordinates on the entrance window of the ionization chamber.

Distinct spatial structures are observed, revealing localized detector imperfections associated 
with the detector geometry. In particular, the correlations observed along the horizontal direction 
coincide with the positions of the window-support wires, indicated by the red arrows 
in Fig.~\ref{fig:detectorimperfections}(b), while the vertical window deformation is observed in Fig.~\ref{fig:detectorimperfections}(a).

These results demonstrate that the proposed framework naturally separates global detector 
calibration from localized detector defects.

Figure~\ref{fig:detectorimperfections}(c) shows the temporal evolution of selected calibration coefficients evaluated every $12$ hours during data taking.
Monitoring the calibration coefficients as functions of time enables the identification of gradual 
changes in detector response. The calibration coefficients therefore constitute 
detector-state observables that can be monitored continuously to assess detector health and performance.

This capability extends the role of artificial intelligence beyond data analysis toward self-monitoring 
instrumentation and predictive maintenance.

\subsection{Atomic Number Identification}

The ultimate objective of charge-state reconstruction is to enable unambiguous particle identification.

Figure~\ref{fig:atomicnumber} presents the two-dimensional correlation between the energy loss 
and residual energy reconstructed using the calibration coefficients inferred by the proposed framework.
The resulting identification chart exhibits well-separated bands corresponding to different 
atomic numbers, demonstrating that the automatically determined calibration coefficients provide 
sufficient accuracy for downstream analysis tasks. This excellent separation enables the autonomous 
generation of high-fidelity labels for atomic number ($Z$) identification, completing a fully 
self-supervised pipeline for comprehensive particle identification.

These results confirm that the proposed self-supervised calibration framework not only reconstructs 
the ionic charge state but also enables subsequent identification of nuclear species.

More generally, this example illustrates how physics-informed self-supervised calibration can provide 
the foundation for complete analysis pipelines in which detector calibration, label generation, and physics 
inference are integrated within a unified framework.

Taken together, these results demonstrate that the proposed framework simultaneously provides accurate 
detector calibration, automatic label generation, and quantitative detector-state observables, establishing 
a pathway toward self-calibrating and self-monitoring scientific instrumentation.

\section{Discussion}

The present work demonstrates that detector calibration can be reformulated as a physics-informed 
self-supervised learning problem. Rather than treating calibration as a prerequisite for data analysis, 
the proposed framework integrates calibration directly into the learning process through physically 
motivated consistency constraints.

This approach addresses a common challenge across scientific instrumentation: labelled datasets 
are often unavailable until extensive calibration procedures have been completed. By exploiting 
known physical properties of the measured system, the proposed framework reduces dependence 
on manual calibration and expert-generated labels.

Traditionally, calibration parameters are viewed as intermediate quantities required for data analysis. 
An important outcome of this work is that they become meaningful detector-state observables.
Their temporal evolution provides direct information about changes in detector performance.

The comparison with the specialized supervised network highlights the complementary roles 
of self-supervised and supervised approaches. The former provides absolute calibration and 
automatic label generation, whereas the latter can account for localized detector effects and 
higher-order corrections.

Although demonstrated using the VAMOS++ magnetic spectrometer, the proposed methodology 
is applicable to a broader class of scientific instruments characterized by latent calibration parameters, 
limited labelled data, and known physical constraints.

Several limitations should nevertheless be considered. The proposed framework assumes the 
existence of informative physical constraints and may require weak prior information to initialize 
the optimization. In addition, the convergence properties of iterative pseudo-labelling procedures 
warrant further investigation.

Future work will focus on extending the framework to additional detector subsystems and 
integrating self-supervised calibration with online monitoring and adaptive control strategies.

\section{Conclusions}

We have presented a physics-informed self-supervised framework that jointly estimates detector 
calibration parameters and ionic charge states directly from raw detector measurements.

Applied to the VAMOS++ magnetic spectrometer, the proposed method successfully reconstructs 
high-quality ionic charge-state and mass distributions without requiring pre-calibrated detector 
signals or externally provided labels.

Beyond automatic calibration, the inferred calibration coefficients provide detector-state observables 
that enable continuous monitoring of detector performance and facilitate the identification of temporal 
drifts and localized detector imperfections.

More generally, this work demonstrates how physical knowledge can replace manual annotation as a 
source of supervision for scientific machine learning. By integrating calibration, monitoring, and 
analysis within a unified framework, the proposed approach establishes a pathway toward 
intelligent instrumentation and, ultimately, autonomous scientific experiments.

This work demonstrates a universal methodology for any high-throughput physical array where manual 
labeling is impossible due to changing environmental drifts, but where the global sensor array is 
fundamentally bound by quantized, discrete physical laws.

\bibliographystyle{cas-model2-names}

\bibliography{Bib}

\end{document}